\documentclass[letterpaper]{article} 
\usepackage{aaai2026}  
\usepackage{times}  
\usepackage{helvet}  
\usepackage{courier}  
\usepackage[hyphens]{url}  
\usepackage{graphicx} 
\usepackage{natbib}  
\usepackage{caption} 
\usepackage{algorithm}
\usepackage{algorithmic}
\usepackage{float}
\usepackage{multirow}
\usepackage{color}
\usepackage{gensymb}
\usepackage{amsmath}

\usepackage{xspace}

\usepackage{graphicx}

\usepackage{amssymb}
\usepackage{booktabs}

\usepackage{colortbl}
\definecolor{yyellow}{rgb}{1, 1, 0.7}
\definecolor{oorange}{rgb}{1, 0.85, 0.7}
\definecolor{rred}{rgb}{1, 0.7, 0.7}

\usepackage{newfloat}
\usepackage{listings}
\DeclareCaptionStyle{ruled}{labelfont=normalfont,labelsep=colon,strut=off} 
\floatstyle{ruled}
\newfloat{listing}{tb}{lst}{}
\floatname{listing}{Listing}
\title{Splat-SAP: Feed-Forward Gaussian Splatting for Human-Centered Scene with Scale-Aware Point Map Reconstruction}
\author {
    Boyao Zhou\textsuperscript{\rm 2,1},
    Shunyuan Zheng\textsuperscript{\rm 2},
    Zhanfeng Liao\textsuperscript{\rm 1},
    Zihan Ma\textsuperscript{\rm 1}\thanks{Work done during an internship at Tsinghua University.},
    Hanzhang Tu\textsuperscript{\rm 1},\\
    Boning Liu\textsuperscript{\rm 1},
    Yebin Liu\textsuperscript{\rm 1}\thanks{Corresponding author (liuyebin@mail.tsinghua.edu.cn).}
}
\affiliations{
    \textsuperscript{\rm 1}Department of Automation, Tsinghua University\\
    \textsuperscript{\rm 2}Ant Group

}

\usepackage{bibentry}

\begin{document}
\maketitle

\begin{abstract}
We present Splat-SAP, a feed-forward approach to render novel views of human-centered scenes from binocular cameras with large sparsity.
Gaussian Splatting has shown its promising potential in rendering tasks, but it typically necessitates per-scene optimization with dense input views.
Although some recent approaches achieve feed-forward Gaussian Splatting rendering through geometry priors obtained by multi-view stereo, such approaches still require largely overlapped input views to establish the geometry prior.
To bridge this gap, we leverage pixel-wise point map reconstruction to represent geometry which is robust to large sparsity for its independent view modeling.
In general, we propose a two-stage learning strategy.
In stage 1, we transform the point map into real space via an iterative affinity learning process, which facilitates camera control in the following.
In stage 2, we project point maps of two input views onto the target view plane and refine such geometry via stereo matching. 
Furthermore, we anchor Gaussian primitives on this refined plane in order to render high-quality images.
As a metric representation, the scale-aware point map in stage 1 is trained in a self-supervised manner without 3D supervision and stage 2 is supervised with photo-metric loss.
We collect multi-view human-centered data and demonstrate that our method improves both the stability of point map reconstruction and the visual quality of free-viewpoint rendering.
Our project page is available at {\small\url{https://yaourtb.github.io/Splat-SAP}}.
\end{abstract}    
\section{Introduction}
\label{sec:intro}

\begin{figure}[h!]
  \centering
  \includegraphics[width=1.0\linewidth]{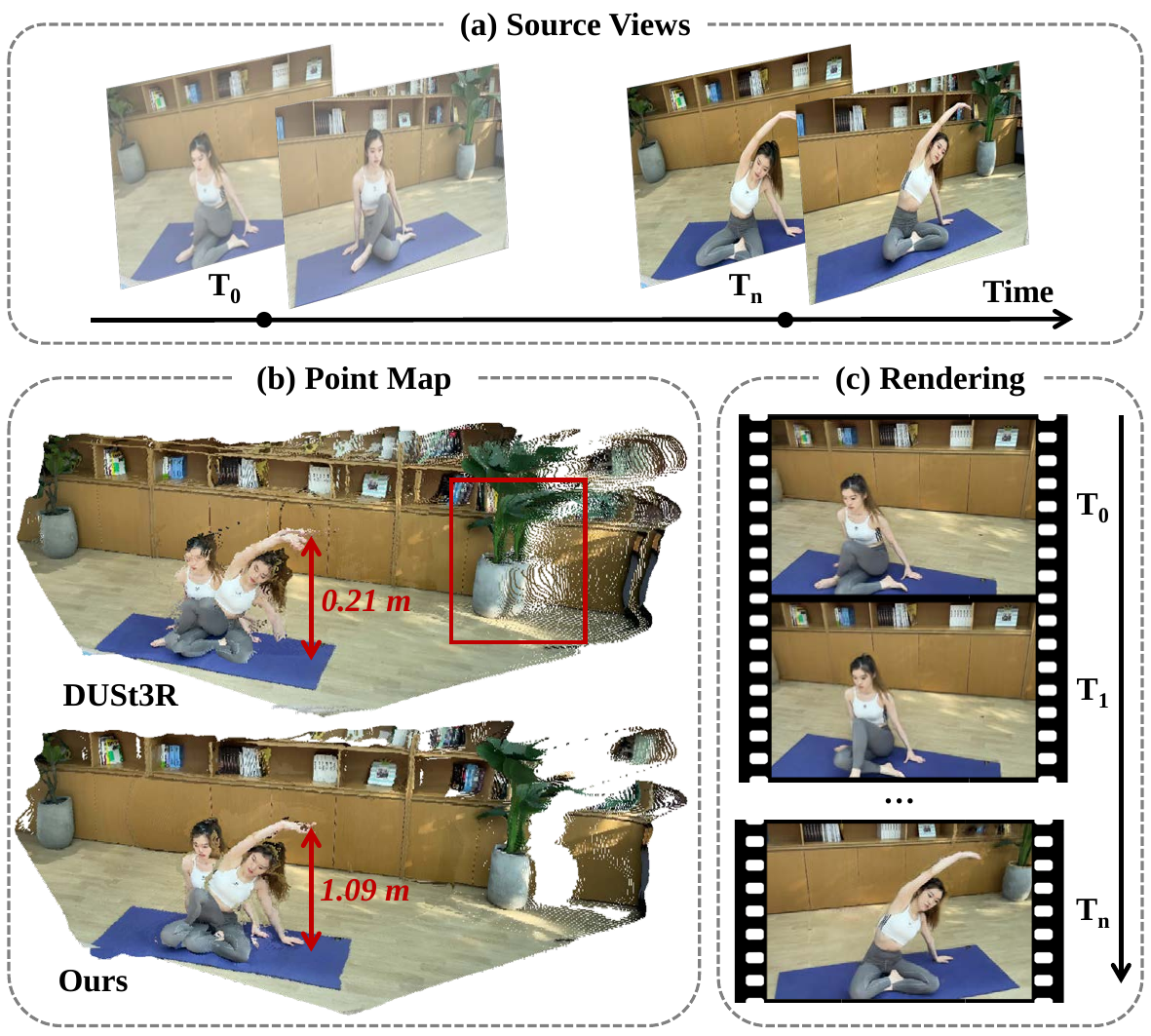}
  \vspace{-6mm}
  \caption{\textbf{Human-centered scene reconstruction and free-view video synthesis.} (a) Source view inputs, (b) our metric scale point map reconstruction, and (c) free-view rendering with our feed-forward Gaussian Splatting.}
  \vspace{-6mm}
  \label{fig:teaser}
\end{figure}

Feed-forward free-viewpoint video synthesis is a crucial task, especially in the setting of sparse views, which could serve many downstream applications such as telecommunications, stage/sports broadcasts, and so on.
Existing pipelines are typically based on differentiable rendering~\cite{mildenhall2020nerf, xu2022point-nerf, yu2021pixelnerf, chen2021mvsnerf, lin2022enerf, Wu2024_4dgs, sun20243dgstream} with the development of neural network. 
Particularly, Gaussian Splatting~\cite{kerbl2023_3dgs} shows its advancement for the high efficiency of rendering and the capable mechanism of back-propagation, but relies on minute-level optimization for each scene and very dense input views.

Recently, Gaussian related methods~\cite{chen2024mvsplat, liu2024mvsgaussian, charatan2023pixelsplat} achieve instant inference in a feed-forward manner, avoiding per-scene optimization, for real-time applications such as telecommunication systems~\cite{tu2024tele} and human-scene synthesis~\cite{zheng2024gpsgaussian, zhou2024gps}.
However, a common strategy is to estimate Gaussian primitive maps defined on the source views, leveraging geometry proxies of multi-view stereo (MVS)~\cite{chen2024mvsplat, liu2024mvsgaussian} or binocular stereo-matching~\cite{zheng2024gpsgaussian, zhou2024gps}.
Such methods require a large overlap of paired images, which increases the redundancy of Gaussian primitives in this overlap area. 
Otherwise, they could not provide reasonable geometry prior when input cameras are with large sparsity.

More recently, DUSt3R~\cite{wang2024dust3r, leroy2024grounding} proposes a novel geometry representation as point maps of input views, which assigns each pixel of input images to a free 3D point.
Unlike traditional MVS methods~\cite{yao2018mvsnet, yang2020cost}, DUSt3R gets rid of stereo constraints and achieves pixel-aligned point maps of binocular inputs from very sparse views, by training on immense 3D geometry data.
To alleviate the impact of the high degree of freedom, DUSt3R and its follow-ups~\cite{leroy2024grounding, wang2024moge} normalize the scale of the reconstructed point map with an average point distance of each scene, which typically causes a dramatic instability of reconstruction in consecutive frames, see Fig.~\ref{fig:teaser}(b). 
Some recent methods~\cite{smart2024splatt3r, ye2024nopose} leverage point map representation for static scene rendering in canonical space.
However, human movement in the scene would cause a relative depth difference in canonical space and lead to large jitters for free-view video rendering, due to the lack of stereo constraint.
In addition, training a foundation model for scale-aware geometry typically requires immense 3D data, but it is always tedious and cumbersome to acquire 3D geometry data.
Therefore, the key point is to obtain scale-aware geometry in a self-supervised manner and to ease the burden of 3D data acquisition.

In this paper, we propose Splat-SAP to achieve human-centered scene reconstruction in real metric space and feed-forward rendering of novel views via Gaussian plane, when inputting a pair of images and camera calibration.
Unlike \cite{wang2024dust3r, leroy2024grounding} representing a scale-invariant point map in canonical space, we inject camera intrinsic embedding~\cite{ye2024nopose} and global image feature~\cite{wang2025vggt} into a network as input to learn a \textbf{scaling} factor to transform the estimated point map from canonical space to real space.
Since the point representation in the original design of DUSt3R~\cite{wang2024dust3r, leroy2024grounding} lacks stereo constraints between two source views, there always exists misalignment between the two point maps.
Thus, we compute the cost between 2 source views to do an iterative \textbf{\textit{coarse registration}} of 2 reconstructed point maps, by projecting the feature from one view to another with calibrated camera pose.
This registration is in the format of a translation map, denoting pixel-wise shift.
Our scaling factor by intrinsic embedding and translation learning by extrinsic projection compose exactly an \textbf{affine transformation} of point map from canonical space to real space.

In terms of rendering, we anchor Gaussian primitives directly on the target view as a Gaussian plane, so that reducing the redundancy of using directly two point maps of source views as Gaussian positions~\cite{zheng2024gpsgaussian, chen2024mvsplat, smart2024splatt3r, ye2024nopose}.
Depth of the Gaussian plane is initialized by projecting two point maps via $\alpha$-blending~\cite{kerbl2023_3dgs}, which largely eases the burden of accurate depth estimation.
Further, we do a \textbf{\textit{fine registration}} with strict stereo constraint~\cite{yao2018mvsnet, lin2022enerf, liu2024mvsgaussian}, which relies on a 3D cost volume by sampling several depth candidates along each pixel ray near the initialized depth. 
With such 3D cost representation, we can aggregate more 3D information to overcome the unobservation issue due to the large sparsity, and to estimate accurate depth.  
The color of the Gaussian plane can be initialized by warping source view pixels directly via the estimated depth.
Additionally, we incorporate both fine 2D and dense 3D features to estimate Gaussian primitives and refine Gaussian color for high-quality rendering.

More importantly, our pipeline can be trained without geometry supervision, different from \cite{wang2024dust3r, leroy2024grounding, wang2024moge}.
To this end, we collect large-scale multi-view data of over 10,000 frames of motion sequences of human-centered scenes to train our model.
We validate the effectiveness of our method on diverse camera settings, \textit{e.g.} industry camera, mobile phone, and GoPro, for both reconstruction and rendering tasks.
In summary, we claim three following contributions:
\begin{itemize}
    \item We introduce a feed-forward pipeline to reconstruct scale-aware point maps and to render free-view video of human-centered scenes, where the point maps are trained in a self-supervised manner without any 3D supervision. 
    \item We propose a 2D coarse to 3D fine registration strategy to estimate scale-aware point maps with a learnable affinity.
    \item We design a Gaussian plane, leveraging scale-aware point maps and incorporating both 2D and 3D features, to guarantee the efficiency and completeness of rendering.
\end{itemize}

\begin{figure*}[h!]
  \centering
  \includegraphics[width=1\textwidth]{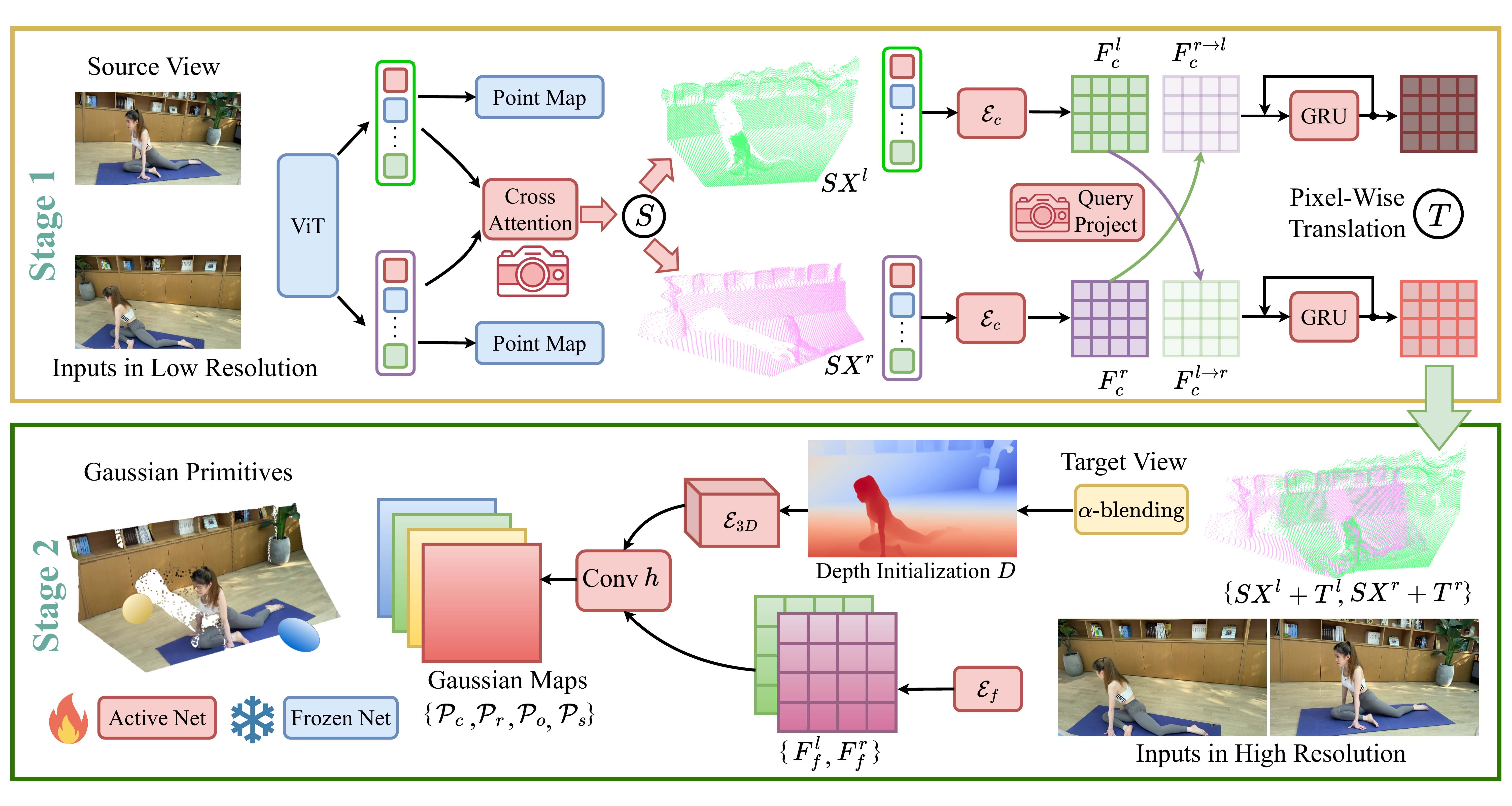}
  \vspace{-6mm}
  \caption{\textbf{Overview of Splat-SAP.} Our method consists of two stages. In the first stage, we take two coarse images as input and predict corresponding point maps, along with an affine transform. In the second stage, our refinement module takes transformed points and fine-resolution images as input, and predicts Gaussian plane of target view for high-quality rendering. }
  \vspace{-4mm}
  \label{fig:pipeline}
\end{figure*}

\section{Related Work}

\subsection{Novel View Synthesis}

Neural Radiance Fields~\cite{mildenhall2020nerf,barron2021mip} achieve photo-realistic rendering quality by applying volume rendering which aggregates the sampled neural features along the ray. 
Recently, 3D Gaussian Splatting~\cite{kerbl2023_3dgs} has made significant advances in neural rendering for its real-time rendering efficiency.
This outstanding technique models static scenes by optimizing a set of Gaussian primitives~\cite{kerbl2023_3dgs, lu2024scaffold, yu2024mip}, including properties of position, scaling, rotation, and opacity. 
Some methods progress further to model dynamic scenes with time-varying Gaussian primitives~\cite{li2024spacetime, yang2024deformable, yan2024street, sun2025splatter}, with 4D neural representation~\cite{Wu2024_4dgs, xu20244k4d, xu2024representing}, or with on-the-fly streamable training~\cite{luiten2023dynamic, sun20243dgstream, girish2025queen, gao2025hicom}.
Although immensely accelerated, the inevitable per-scene optimization still requires minutes to accomplish.

To eliminate the long-time optimization process, generalizable neural rendering methods~\cite{wang2021ibrnet, yu2021pixelnerf, charatan2023pixelsplat, zheng2024gpsgaussian} have been developed for feed-forward novel view rendering.
Typically, these generalizable paradigms resort to leveraging the learned 2D/3D priors from extensive data~\cite{zhou2018stereo, liu2021infinite, yu2021function4d} to ease the long-term optimization.
In this practice, ENeRF~\cite{lin2022enerf} integrates cost volume to provide a coarse depth initialization and thus reduce the sampling points, leading to an efficient framework.
With respect to Gaussian-based methods~\cite{xu2024depthsplat, charatan2023pixelsplat, zheng2024gpsgaussian, liu2024mvsgaussian}, the solution is parallel to that in generalizable NeRF, \textit{e.g.} by using epipolar stereo and cost volume.
However, some methods~\cite{zheng2024gpsgaussian} necessitate ground truth depth for training, and others~\cite{charatan2023pixelsplat, chen2024mvsplat, zhou2024gps} are limited under sparse input views, due to the difficulty of establishing correlation with small overlap of input views.

\subsection{3D Reconstruction}

Multi-view stereo is a traditional 3D reconstruction technique, which can be categorized according to the output modality, including point cloud~\cite{lhuillier2005quasi, furukawa2009accurate}, volumetric representation~\cite{seitz1999photorealistic, kutulakos2000theory}, and depth map~\cite{campbell2008using, schonberger2016pixelwise}.
MVSNet~\cite{yao2018mvsnet} opens up the era of deep learning-based MVS methods.
Binocular Stereo~\cite{zabih1994non}, as a special kind of MVS, aims to find the maximum correspondence on the horizontal epipolar line.
However, such methods struggle with invisible issues and large sparsity of input views.
Progressing further, the neural implicit surface methods~\cite{wang2021neus, wang2023neus2, li2023neuralangelo}, as a variant of neural radiance fields, perform accurate 3D reconstruction with only rendering loss, which avoids collecting 3D properties for training.
With the prevalence of 3D Gaussian Splatting in NVS, a body of research~\cite{huang20242d, dai2024high, guedon2024sugar, lyu20243dgsr} attempts to adapt it to multi-view 3D Reconstruction with flattened or surfel shaped Gaussian primitives.
These neural rendering based methods typically rely on dense input views to supervise with rendering loss and long-time optimization.

More recently, DUSt3R\cite{wang2024dust3r} proposes a novel 3D representation, defining point maps on a pair of source views, aligning a pixel to a free 3D point, bypassing the need for camera poses.
In such ill-posed conditions, both DUSt3R and its follow-ups~\cite{leroy2024grounding, wang2024moge} bound point maps in scale-invariant canonical space.
NoPoSplat~\cite{ye2024nopose} and Splat3R~\cite{smart2024splatt3r} incorporate such representation into the rendering pipeline for static scenes.
Some followers~\cite{lu2024align3r, zhang2024monst3r} point out that when handling dynamic scenes, DUSt3R encounters two limitations: (1) the misaligned background points, and (2) incorrect foreground depth estimation, causing some regions placed in the background.
They address these issues with a global test time optimization on the whole video, while we commit to probing a feed-forward solution in this paper.

\section{Method}
\label{sec:method}

Given a pair of images and camera calibration, our method reconstructs scale-aware point maps with an affinity learning in the first stage.
In the second stage, we project such point maps onto the target view and refine this geometry to anchor Gaussian primitives for the rendering task. 
An overview of our 2D-coarse-to-3D-fine pipeline is shown in Fig.~\ref{fig:pipeline}. 

\subsection{Scale-Aware Geometry Reconstruction}
\label{sec:geometry}
\paragraph{Point Map.} It is introduced by DUSt3R~\cite{wang2024dust3r} as a novel but scale-invariant representation $X \in [0, 1]^{W \times H \times 3}$ of 3D scene, which is associated with corresponding image $I$ of resolution $W \times H$. 
We apply the follow-up, MASt3R~\cite{leroy2024grounding}, to predict two pieces of point map from source views, $i \in \{l, r\}$ for left and right view, to represent the scene.
As a coarse stage, we take $W = 512$ and $H=288$.
MASt3R encodes the input images into patches of features $F^{i}$ with ViT and then decodes them into $X^i$ in canonical space.
Without any stereo constraint, it is capable of predicting reasonable geometries from two cameras in a large sparsity, but hard to control the target view with real camera parameters, leading to jitters when inferring consecutive frames.
In the following, we transform point maps from canonical space to real space in an absolute scale with an affinity, in the format of scaling $S \in \mathbb{R}^3$ and translation $T \in \mathbb{R}^{W \times H \times 3}$.

\paragraph{Scaling.} It is a global factor related to camera intrinsic parameters such as focal $f$. In addition, the distance $d$ between two cameras provides a cue of measurement in real space. Thus, we embed them with positional encoding~\cite{mildenhall2020nerf} $\text{PE}$
\begin{equation}
    e = \text{PE}(f, d)
\end{equation}
We further process the encoded features $F$ with self- $Att_s$ and cross-attention $Att_c$
\begin{align}
\begin{split}
\label{formula:att}
    \langle\mathbf{Q,K,V}\rangle &= \langle F \mathbf{W}^Q, F \mathbf{W}^K, F \mathbf{W}^V \rangle \\
    f_s &= \text{Avg}(Att_s(\mathbf{Q}_l, \mathbf{K}_l, \mathbf{V}_l)) \\
    f_c &= \text{Avg}(Att_c(\mathbf{Q}_l, \mathbf{K}_r, \mathbf{V}_r))
\end{split}
\end{align}
where the average operator is used to extract global information. We use an $MLP$ to compute the scaling factor
\begin{equation}
    S = MLP(f_s, f_c, e)
    \label{eq:scale}
\end{equation}
We note that the degree of freedom of $S$ is 3 to deal with the distortion of original point map reconstruction of MASt3R.

\paragraph{Translation.} Although the point maps could be rescaled to real space by multiplying the scaling factor, there could still exist point-wise shifts due to the lack of stereo constraint in MASt3R. 
Inspired by the view consistency check~\cite{yan2020dense} in MVS, we believe that such a shift depends not only on the features in one view but also on the corresponding features in the other view. 
Thus, we process the aforementioned ViT features with a lightweight convolutional encoder $\mathcal{E}_c$ to yield feature map $F_c^i = \mathcal{E}(F^i)$. Furthermore, we obtain the feature map $F^{j \rightarrow i}_c$ in the other view $j$, $j \in \{ l, r \}$ and $i \neq j$, by first projecting the rescaled points $SX^i$ onto the view $j$ and then querying the corresponding features with bilinear sampling
\begin{equation}
    F^{j \rightarrow i}_c = \text{Query}(F^j_c , \text{Proj}(SX^i, K^j))
\end{equation}
where $K$ is the camera parameter.
We follow the idea of iterative updating~\cite{teed2020raft,lipson2021raft-stereo} to compute point-wise translation with $\text{GRU}$ operator~\cite{cho2014properties} by considering feature maps from both views and the position of each point
\begin{equation}
    T^i = \text{GRU}(F^i, F^{j \rightarrow i}, SX^i)
\end{equation}
We obtain the position of the point set in real space with the learned affine transform
\begin{equation}
    X_{t}^i = SX^i + T^i
\end{equation}

\subsection{Rendering via Gaussian Plane}
\label{sec:rendering}

\paragraph{3D Refinement.} When splatting the aforementioned point set to the target view, we still observe inevitable jitters and holes due to the lack of 3D stereo constraint, see Fig.~\ref{fig:abla_full}(a). 
Given the paired images $I_f$ in fine resolution of $W = 1024$ and $H=576$, we encode them with convolutional layers $\mathcal{E}_f$ into $F_f^i = \mathcal{E}_f(I_f^i)$. 
Additionally, we project the transformed point set $X_t$ onto the target view $k$ with $\alpha$-blending mechanism in Gaussian Splatting to yield the initial depth map $\mathcal{D}^k$.
For each pixel $(u,v)$, we sample several position candidates $\{d_1, d_2,...d_N\}$ near the initial depth value $d = \mathcal{D}(u,v)$ along camera ray.
For each candidate $d(u,v,n)$, we warp the feature from the source views to the target view
\begin{equation}
    p^k(u,v,n) = \text{Proj}^{-1}(d^k(u,v,n), K^k)
\end{equation}
\begin{equation}
\label{eq:warp}
    F_f^{i \rightarrow k}(u,v,n) = \text{Query}(F_f^i , \text{Proj}(p^k(u,v,n), K^i))
\end{equation}
where the warping process can be efficiently achieved by matrix operation.
We further process the aggregation of features with 3D convolutions $\mathcal{E}_{3D}$ into a feature volume 
\begin{equation}
    \Phi^k = \mathcal{E}_{3D}(F_f^{l \rightarrow k} , F_f^{r \rightarrow k})
\end{equation}
Following ENeRF~\cite{lin2022enerf}, we compute the depth probability distribution $w_n$ along the camera ray by regressing with the feature volume $\Phi$.
The final position of Gaussian primitives can be represented with the refined depth $\bar{d} = \Sigma_n w_n d_n$.

\paragraph{Gaussian Plane.} Once the position of Gaussian is determined, Gaussian plane $\mathcal{G}$ consists of four attribute maps of color, rotation, scaling and opacity
\begin{equation}
    \mathcal{G} = \{ \mathcal{P}_c, \mathcal{P}_r, \mathcal{P}_s, \mathcal{P}_o \}
\end{equation}
Using the warping process in Eq.~\ref{eq:warp}, we obtain color $\{ C^{l \rightarrow k}, C^{r \rightarrow k} \}$ of target view warped from source views.
We further query the feature $\phi$ from feature volume $\Phi$ for Gaussian primitives via tri-linear interpolation.
We thus learn a weighted color to initialize the Gaussian color
\begin{align}
    w^{i}_{c} &= MLP_{c}(f_f^l, f_f^r, \phi)\\
    C^k &= \Sigma_{i} w_c^i C^{i \rightarrow k} 
    \label{eq:color_stage2}
\end{align}
We arrange all features into the format of a 2D map and further aggregate them into the feature map $\mathcal{M} = \text{Agg}\{ f_f^l, f_f^r, \phi \}$.
Following GPS-Gaussian, we yield rotation, scaling, and opacity map via convolutional heads $h_a, a=\{r, s, o \}$, considering the Gaussian position $Y$
\begin{equation}
    \mathcal{P}_a = h_a (\mathcal{M}, Y)
\label{eq:ha}
\end{equation}
In addition, we update the initial color with a learned residual color map 
\begin{align}
    \Delta C &= {h}_c(\mathcal{M}, Y, C)\\
    \mathcal{P}_c &= \alpha C + (1-\alpha) \Delta C
    \label{eq:color}
\end{align}
Finally, we splat the Gaussian plane $\mathcal{G}$ in a fine resolution of $1024 \times 576$ to render an image $\hat{I}$ in a higher resolution of $1280 \times 720$.

\begin{table*}[ht!]
\small

\centering
\begin{tabular}{l|ccc|ccc|ccc}
\toprule
\multicolumn{1}{c}{\multirow{1}[6]{*}{Method}} & \multicolumn{3}{c}{Camera} & \multicolumn{3}{c}{GoPro} & \multicolumn{3}{c}{Mobile} \\

\cmidrule{2-10} \multicolumn{1}{c}{} & PSNR$\uparrow$ & SSIM$\uparrow$ & \multicolumn{1}{c}{LPIPS$\downarrow$} & PSNR$\uparrow$ & SSIM$\uparrow$  & \multicolumn{1}{c}{LPIPS$\downarrow$} & PSNR$\uparrow$ & SSIM$\uparrow$ & \multicolumn{1}{c}{LPIPS$\downarrow$} \\ \midrule

NoPoSplat  & 25.035 & 0.866 & 0.173 & 26.128 & 0.889 & 0.121 & 21.594 & 0.591 & \underline{0.272}  \\
4D-GS  & 27.814 & 0.906 & 0.150 & 27.244 & 0.907 & 0.205 & 25.655 & \underline{0.825} & 0.284  \\
MVSplat     & 27.899 & 0.902 & 0.148 & \underline{29.942} & 0.934 & 0.157 & \textbf{26.545} & 0.805 &  0.314  \\
MVSGaussian & \underline{29.326} & \underline{0.957} & \textbf{0.069} & 27.413 & 0.926 & 0.151 & 19.927 & 0.683 & \underline{0.272}   \\
ENeRF  & 28.272 & 0.943 & 0.084 & 29.906 & \underline{0.943} & \underline{0.108} & 20.579 & 0.640 & 0.302   \\
Ours   & \textbf{32.220} & \textbf{0.957} & \underline{0.079} & \textbf{31.640} & \textbf{0.955} & \textbf{0.096} & \underline{25.721} & \textbf{0.827} & \textbf{0.244} \\
\bottomrule
\end{tabular}
\vspace{-1mm}
\caption{\textbf{Quantitative comparison of rendering methods on multi-view datasets.} NoPoSplat~\cite{ye2024nopose}, MVSplat~\cite{chen2024mvsplat} and MVSGaussian~\cite{liu2024mvsgaussian} are feed-forward Gaussian Splatting methods and ENeRF~\cite{lin2022enerf} is feed-forward NeRF based method, while 4D-GS~\cite{Wu2024_4dgs} is optimization based 4D Gaussian Splatting method. \textbf{Bold} highlights the top-performing method, while \underline{underline} indicates suboptimal performance across various evaluation criteria.}
\vspace{-2mm}
\label{tab:num_compare_bg}
\end{table*}

\begin{figure*}[ht!]
  \centering
  \includegraphics[width=0.88\textwidth]{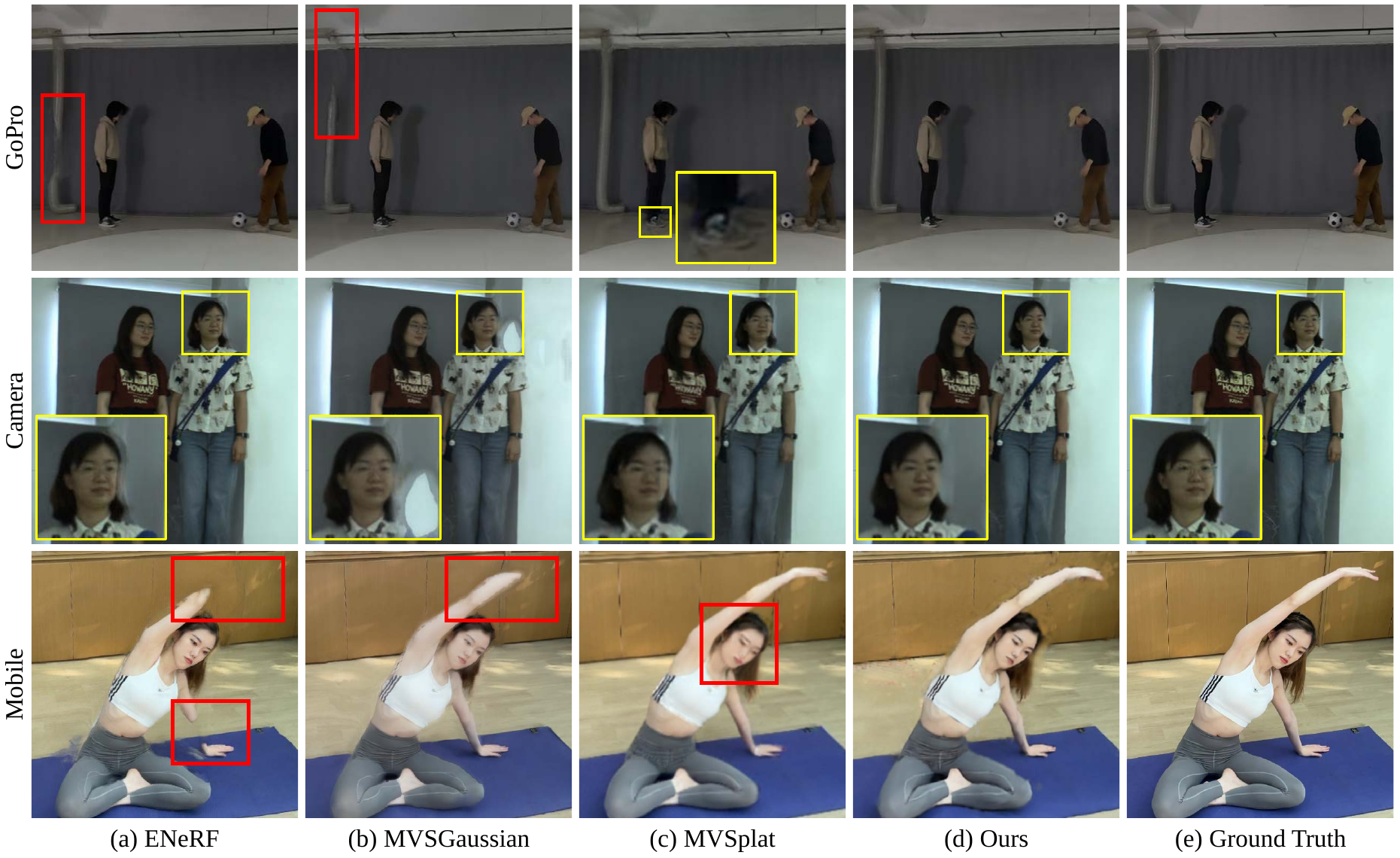}
  \vspace{-2mm}
  \caption{\textbf{Quantitative comparison of rendering.} We show results of (a) ENeRF~\cite{lin2022enerf}, (b) MVSGaussian~\cite{liu2024mvsgaussian}, (c) MVSplat~\cite{chen2024mvsplat}, (d) Ours and (e) Ground Truth for GoPro, Camera and Mobile datasets.}
  \vspace{-4mm}
  \label{fig:quantative_comp}
\end{figure*}

\subsection{Training}

We define the rendering loss as a combination of L1 loss $\mathcal{L}_1$ and SSIM loss~\cite{wang2004ssim} $\mathcal{L}_{ssim}$
\begin{equation}
    \mathcal{L}_{render}(\hat{I}, I^{gt}) = \beta_1 \mathcal{L}_1 + \beta_2 \mathcal{L}_{ssim}
    \label{eq:render_loss}
\end{equation}
where $\hat{I}$ and $I^{gt}$ stand for rendering image and ground truth image.
Since the transformed geometry largely impacts the rendering module, we propose a 2-stage training strategy.

\paragraph{Stage 1.} We firstly train the affine transformed point maps with captured multi-view images in a self-supervised manner without any 3D supervision.
To this end, we predict Gaussian planes $\mathcal{G}^i = \{ \mathcal{P}_p^i, \mathcal{P}_c^i, \mathcal{P}_r^i, \mathcal{P}_s^i, \mathcal{P}_o^i \}, i=\{ l,r \}$ on source views. Among them, the position and color plane can be obtained with point map $X^i_t$ and input image $I^i$. 
We further use the auxiliary layers $\bar{h}_a$, similar to convolutional operator $h_a$ in Eq.~\ref{eq:ha}, to predict $\mathcal{P}_a^i, a=\{r, s, o\}$.
Once the auxiliary Gaussian planes are done, we can render the image on the target view.  
Inspired by GPS-Gaussian+~\cite{zhou2024gps}, we propose a regularization term as Chamfer distance between two 6-dimensional point sets $P^l, P^r$
\vspace{-2mm}
\begin{align}
\begin{split}
    CD(i \rightarrow j) &= \frac{1}{|{P}^i|}\sum_{p_i \in {P}^i} \min_{p_j \in {P}^j} \|p_i - p_j\|_2 \\ 
    \mathcal{L}_{CD} &= CD(l \rightarrow r) + CD(r \rightarrow l)
\end{split}
\label{eq:geo_regular}
\end{align}
where $p_i$ is pixel-wise point on point map $X_t^i(u,v)$ associated with the corresponding pixel color $I^i(u,v)$.
Such a regularization term allows two pieces of point maps to converge to a better geometry.

Therefore, we supervise the affinity learning and auxiliary layers with rendering loss and the regularization term
\begin{equation}
    \mathcal{L}_{stage1} = \mathcal{L}_{render} + \gamma\mathcal{L}_{CD}
    \label{eq:loss1}
\end{equation}
Note that during the training, we freeze the weight of the MASt3R~\cite{leroy2024grounding} network and no longer require the geometry ground truth. 

\paragraph{Stage 2.} The scale-aware point maps in the previous step allow us to initialize the depth of the target view, which largely improves the stability of the training process in stage 2. Specifically, we train the 3D refinement module and Gaussian planes with photo-metric loss. In practice, we have an initial color plane $\hat{I}_f = C$ in fine resolution and a splatting image $\hat{I}_h$ in a higher resolution.
So the training loss is the combination of two rendering losses 
\begin{equation}
    \mathcal{L}_{stage2} = \lambda_1 \mathcal{L}_{render}(\hat{I}_f, I_{f}^{gt}) + \lambda_2 \mathcal{L}_{render}(\hat{I}_h, I_{h}^{gt})
    \label{eq:loss2}
\end{equation}

For both stages 1 and 2, we do not require 3D geometry supervision, which facilitates the training process on real captured 2D images.

\section{Experiment}
\label{sec:exp}

\subsection{Settings}

\paragraph{Data.} We collect multi-view data from 3 types of cameras, including industry camera (THumanMV~\cite{zhou2024gps}), mobile phone (4K4D~\cite{xu20244k4d} and SelfCap~\cite{xu2024representing}) and our captured GoPro data. 
For training, we take 15 training sequences of industry camera data, 6 sequences of GoPro data, and around 3000 frames from 4K4D dance sequence and SelfCap yoga sequence. 
Compared to public datasets, we capture large scenes accommodating sports movement of multi-person with a portable GoPro system.  
To evaluate our Splat-SAP, we take the sequences of unseen characters or of unseen motions from each dataset.
In particular, we train only one model of our affinity module for stage 1, while we train one refinement module per camera type.
We pick 6 cameras, facing to characters. 
The leftmost and the rightmost cameras are fed into the network as source input views.
The other 4 views are used as supervision during training and to compute metrics during evaluation.

\paragraph{Metrics.} For rendering, the quality of synthesized images is measured with widely used PSNR, SSIM~\cite{wang2004ssim} and LPIPS~\cite{zhang2018lpips}.
We apply Chamfer distance of both directions to evaluate the quality of geometry.
We note that the ground truth point set is reconstructed by using Structure-from-Motion~\cite{schoenberger2016sfm} with all 6 views under a long-time optimization.

\paragraph{Implementation Details.} We employ a two-stage training strategy.
We first train the affinity learning module for $100k$ iterations with full training data.
For each camera type, we further train the rendering module for $60k$ iterations in stage 2.
Our networks can be trained on a single RTX 3090 GPU with 24GB.
We set $\alpha = 0.8$ in Eq.~\ref{eq:color}, $\beta_1 = 0.8,\ \beta_2 = 0.2$ in Eq.~\ref{eq:render_loss}, $\gamma = 0.5$ in Eq.~\ref{eq:loss1}, and $\lambda_1 = 0.5,\ \lambda_2 = 0.5$ in Eq.~\ref{eq:loss2}.
The input to the first stage is in coarse resolution of $W = 512, H=288$, while the fine resolution of $W=1024, H=576$ for the second stage. 
We render the high-resolution image of $W=1280, H=720$ in the end.

\subsection{Results}

\begin{figure}[htpb!]
  \centering
  \includegraphics[width=1.\linewidth]{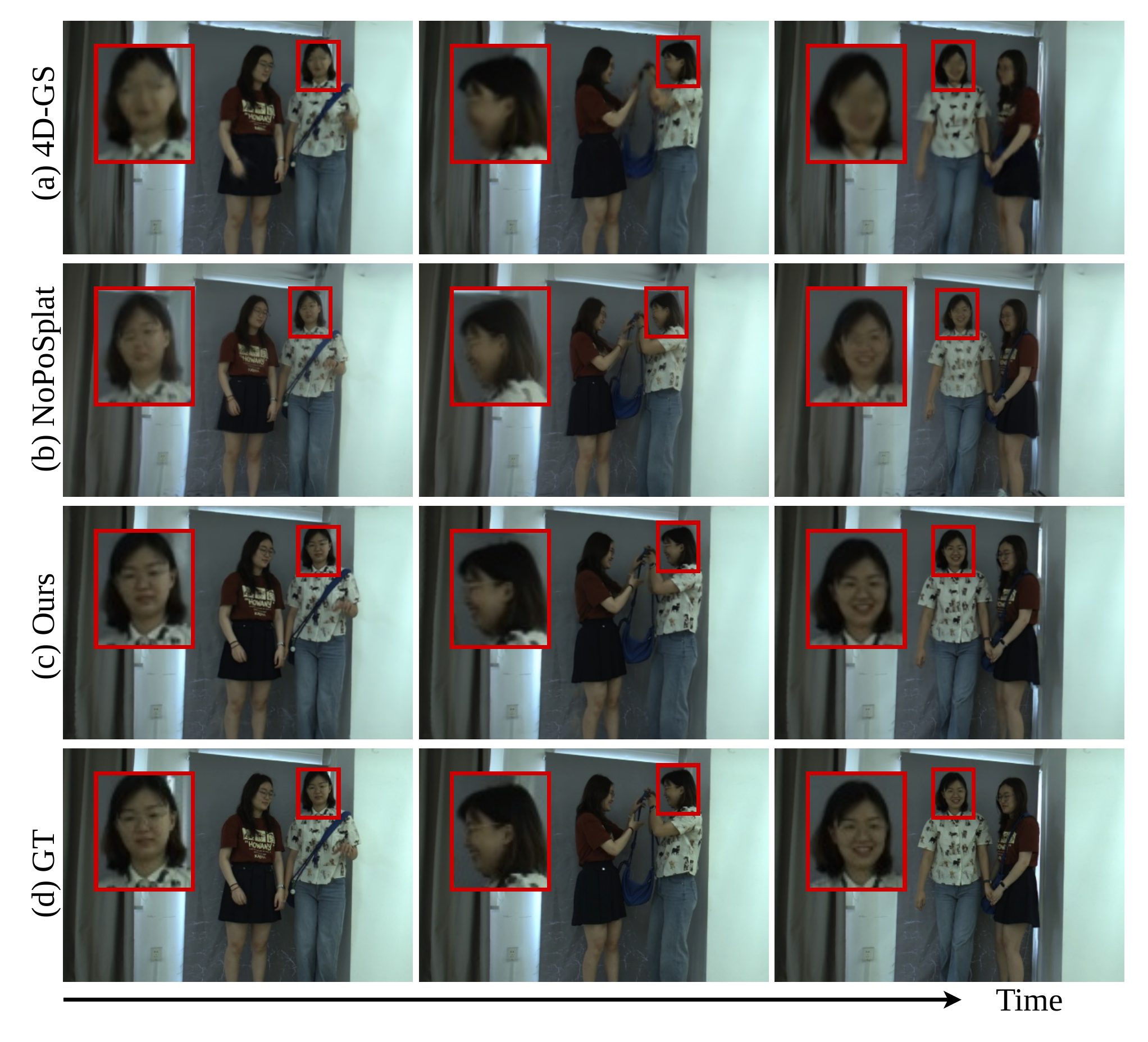}
  \vspace{-8mm}
  \caption{\textbf{Qualitative comparison of rendering on a sequence of data.} Our method preserves temporal and view consistency against 4D-GS and NoPoSplat.}
  \label{fig:compare_noposplat}
  \vspace{-6mm}
\end{figure}

\paragraph{Baselines.} We compare Splat-SAP with state-of-the-art methods of feed-forward rendering, including NeRF-based method ENeRF~\cite{lin2022enerf}, as well as Gaussian-based methods MVSplat~\cite{chen2024mvsplat}, MVSGaussian~\cite{liu2024mvsgaussian} and NoPoSplat~\cite{ye2024nopose}.
In addition, we compare with the optimization-based method 4D-GS~\cite{Wu2024_4dgs}, which requires a long time optimization on sequential data.
We train ENeRF, MVSplat and MVSGaussian from scratch with the same data setting as our second stage training.
We take the pretrained checkpoint of NoPoSplat provided by the original authors and fine-tune it with our training data.
We feed the fine-resolution inputs to ENeRF and MVSGaussian, while the coarse ones to MVSplat and NoPoSplat due to the high memory cost.

For geometry, we compare with scale-invariant methods such as DUSt3R~\cite{wang2024dust3r}, MASt3R~\cite{leroy2024grounding} and VGGT~\cite{wang2025vggt}. 
Furthermore, the scale-aware method Pow3R~\cite{jang2025pow3r} and metric depth estimation method Prompt-DA~\cite{lin2025prompting} are considered for comparison.
Similar to us, Pow3R requires camera calibration as an auxiliary input.
Particularly, Prompt-DA feeds images along with corresponding coarse depth maps to the network as inputs.

\paragraph{Rendering Comparisons.} We report the quantitative results on datasets of Camera, GoPro and Mobile Phone in Tab.~\ref{tab:num_compare_bg}.
Our method, in general, outperforms others, especially on Camera and GoPro datasets.
Since LPIPS~\cite{zhang2018lpips} is sensitive to higher resolution, our rendering in the resolution of $W=1280$ is on par with the results of MVSGaussian in the resolution of $W=1024$ on Camera data.
However, MVSGaussian and ENeRF can not handle thin structures and result in some missing parts in Fig.~\ref{fig:quantative_comp}(a,b), due to the large sparsity.
For mobile data, mobile phones are in the mode of alternate zoom-in and zoom-out, which increases the difficulty for feed-forward methods. 
Although the setting is tough, our method still renders the fine-grained results with respect to MVSplat in Fig.~\ref{fig:quantative_comp}.
Due to the lack of geometry regularization term in Eq.~\ref{eq:geo_regular}, two pieces of Gaussians are sometimes mis-aligned for MVSplat and NoPoSplat in Fig.~\ref{fig:quantative_comp}(c) and \ref{fig:compare_noposplat}(b). 
Thanks to the geometry foundation model MASt3R and our coarse-to-fine learning strategy, we manage the case of sparse-view camera inputs and preserve the view consistency.

NoPoSplat~\cite{ye2024nopose} also leverages MASt3R as a geometry prior, but it gets rid of the traditional stereo constraint, leading to a bad perspective on target view in Fig.~\ref{fig:compare_noposplat}(b). 
During the inference, NoPoSplat still requires source and target view camera poses and normalizes them into a relative scale. 
In Fig.~\ref{fig:compare_noposplat}(b), such normalization leads to rendering jitters in the case of dynamic differences caused by human movement.
4D-GS~\cite{Wu2024_4dgs} also neglects geometric constraint and struggles to achieve temporal consistency, see Fig.~\ref{fig:compare_noposplat}(a), for fast motion under sparse views, even if it optimizes on the sequential data for a long time.

\begin{figure}[t!]
  \centering
  \includegraphics[width=1.\linewidth]{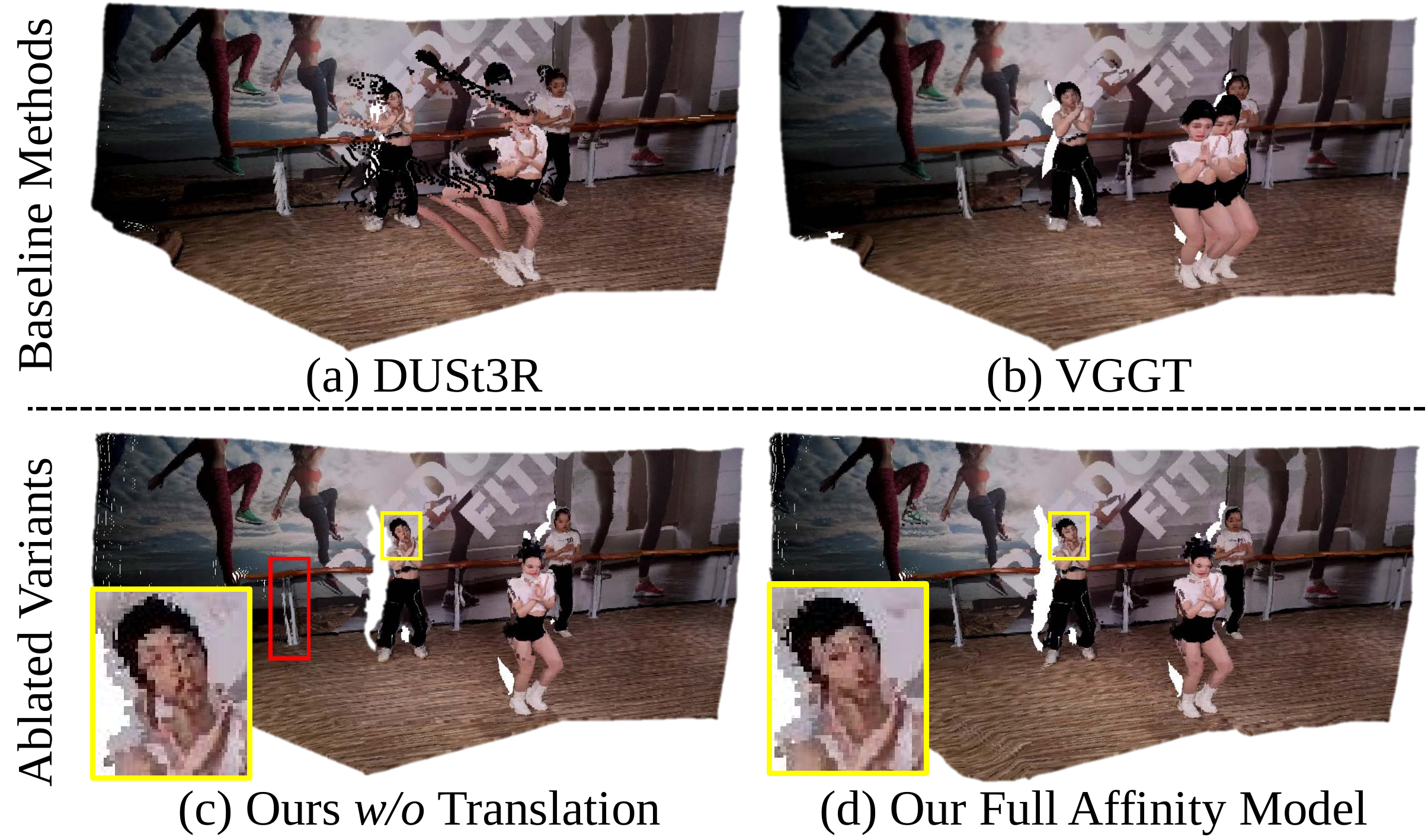}
  \vspace{-6mm}
  \caption{\textbf{Qualitative comparison of geometry.} We show point maps of (a) DUSt3R, (b) VGGT, (c) Ours without pixel-wise translation, and (d) Our full affinity. Here is the point map reconstruction with corresponding pixels.}
  \label{fig:geo_ablation}
  \vspace{-3mm}
\end{figure}

\paragraph{Geometry Comparisons.}

As scale-invariant methods, the original point maps of DUSt3R, MASt3R and VGGT are defined in canonical space.
Therefore, we employ the ground truth scale factor by comparing their bounding box with that of the ground truth to rescale them into real space.
However, DUSt3R sometimes immerses into a local minimum, and projects foreground points onto background, see Fig.~\ref{fig:geo_ablation}(a). 
VGGT is not able to handle two-view input with large sparsity, which leads to misalignment of the foreground human from two input views in Fig.~\ref{fig:geo_ablation}(b).  
Further, the scale of the scene can not be perfectly estimated by Pow3R, even using camera calibration, thus causing a large Chamfer distance in Tab.~\ref{tab:geometry}. 
In addition, we feed the rescaled result of MASt3R as coarse depth input to Prompt-DA.
But the diffusion-based method increases the uncertainty of prediction, and can not preserve 3D consistency from two input views.
Although our method is trained without any geometry loss, we still achieve a superior result in Tab.~\ref{tab:geometry}.
We notice that the ground truth of geometry is a relatively sparse point cloud when using SfM under 6 input views, thus the Chamfer distance from ground truth to prediction can better reflect geometry quality.

\begin{figure}[htpb!]
  \centering
  \includegraphics[width=1.\linewidth]{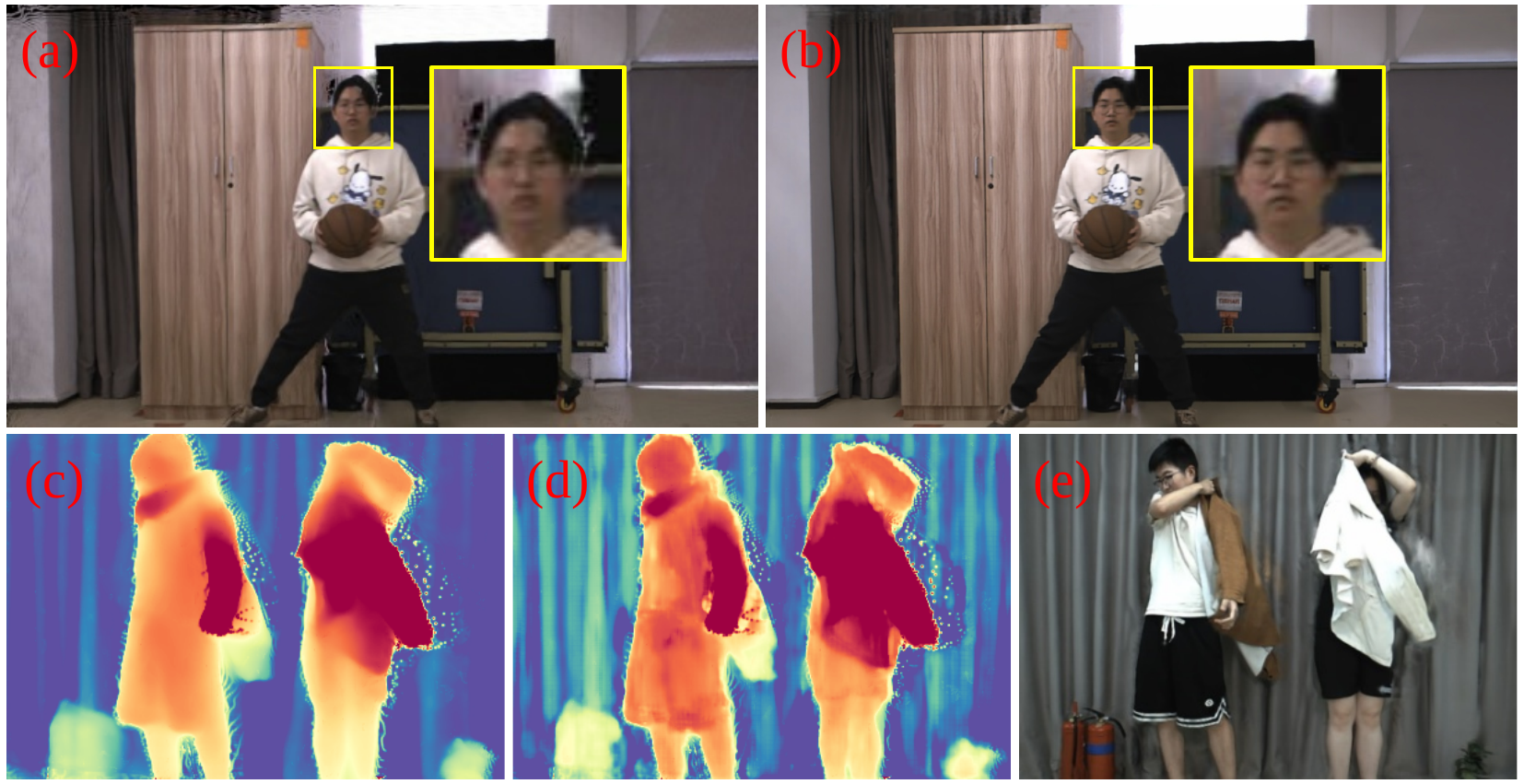}
  \vspace{-4mm}
  \caption{\textbf{Qualitative ablation results.} 
  \textit{Upper row:} the rendering comparison between (a) stage 1 and (b) stage 2.
  \textit{Bottom row} illustrates the effectiveness of our depth refinement module in stage 2: (c) initial depth map rendered by affine transformed point maps, (d) depth map after refinement, and (e) rendering results.}
  \label{fig:abla_full}
  \vspace{-2mm}
\end{figure}

\begin{table}[t!]
\small
\centering
\begin{tabular}{l|cc}
\toprule

\multicolumn{1}{c}{\multirow{1}[0]{*}{Method}} & Pred $\rightarrow$ GT $\downarrow$ & GT $\rightarrow$ Pred $\downarrow$ \\ \midrule

DUSt3R         & 0.305 & 0.160 \\
VGGT           & 0.288 & 0.129 \\
Pow3R          & 0.281 & 0.134 \\
MASt3R         & 0.212 & 0.069 \\
Prompt-DA      & 0.205 & 0.063 \\
\hline 
Ours \textit{w/o} Translation   & 0.191 &  0.046 \\
Our Full Model   & \textbf{0.172}  & \textbf{0.027}  \\

\bottomrule
\end{tabular}
\vspace{-2mm}
\caption{\textbf{Quantitative comparison of geometry.} For scale-invariant methods, we compute the rescale factor by comparing their bounding box with that of the ground truth.}
\vspace{-4mm}
\label{tab:geometry}
\end{table}

\paragraph{Ablation Study.} 
We first evaluate the effectiveness of our pixel-wise translation in stage 1.
Our point maps are rescaled with 3-dimensional scaling factors (Eq.~\ref{eq:scale}) when considering camera intrinsic embedding.
The point maps can still not avoid misalignment by only using the scaling operator, see Fig.~\ref{fig:geo_ablation}(c).
Integrating iterative pixel-wise translation learning, our full affinity module further improves the Chamfer distance in Tab.~\ref{tab:geometry}, and corrects misaligned parts in Fig.~\ref{fig:geo_ablation}(d). 

In addition, we evaluate the effectiveness of our refinement module in stage 2. 
Alternatively, we can directly synthesize the target view with two Gaussian planes learned by auxiliary layers in stage 1.
However, such models typically struggle with holes in the boundary area between foreground and background, see Fig.~\ref{fig:abla_full}(a).
The refinement module can correct artifacts and refine details, see Fig.~\ref{fig:abla_full}(b, d).

\section{Discussion}
\paragraph{Conclusion.} We present Splat-SAP, a feed-forward approach for novel view synthesis of human-centered scenes.
In particular, we employ a 3D foundation model and utilize iterative affinity learning to reconstruct scale-aware point maps as a coarse geometry without 3D supervision.
We further leverage geometric constraints to refine the initial geometry, on which we build a Gaussian plane for rendering.
The full coarse-to-fine pipeline can be trained with only rendering loss by using multi-view image datasets.
Our method achieves superior rendering results with respect to baseline methods, especially in the case of sparse input views.

\paragraph{Limitation.} We notice the floating artifacts in Fig.~\ref{fig:abla_full}(e).
This is because MASt3R might predict some floating points on the boundary between foreground and background.
Since such regions are observed by only one of two input views, they can not be corrected by our refinement module.
We believe that incorporating monocular prior~\cite{xu2024depthsplat} would alleviate this problem.

\paragraph{Acknowledgement.} This paper is supported by National
Key R\&D Program of China (2022YFF0902200) and NSFC
project (Grants 62125107 and 62301298).

\bibliography{aaai2026}

\section{Appendix}

In the supplement, we present more results on geometry, ablation study on rendering, time analysis of our pipeline and data explanation in the following.

\begin{figure}[h!]
  \centering
  \includegraphics[width=1.\linewidth]{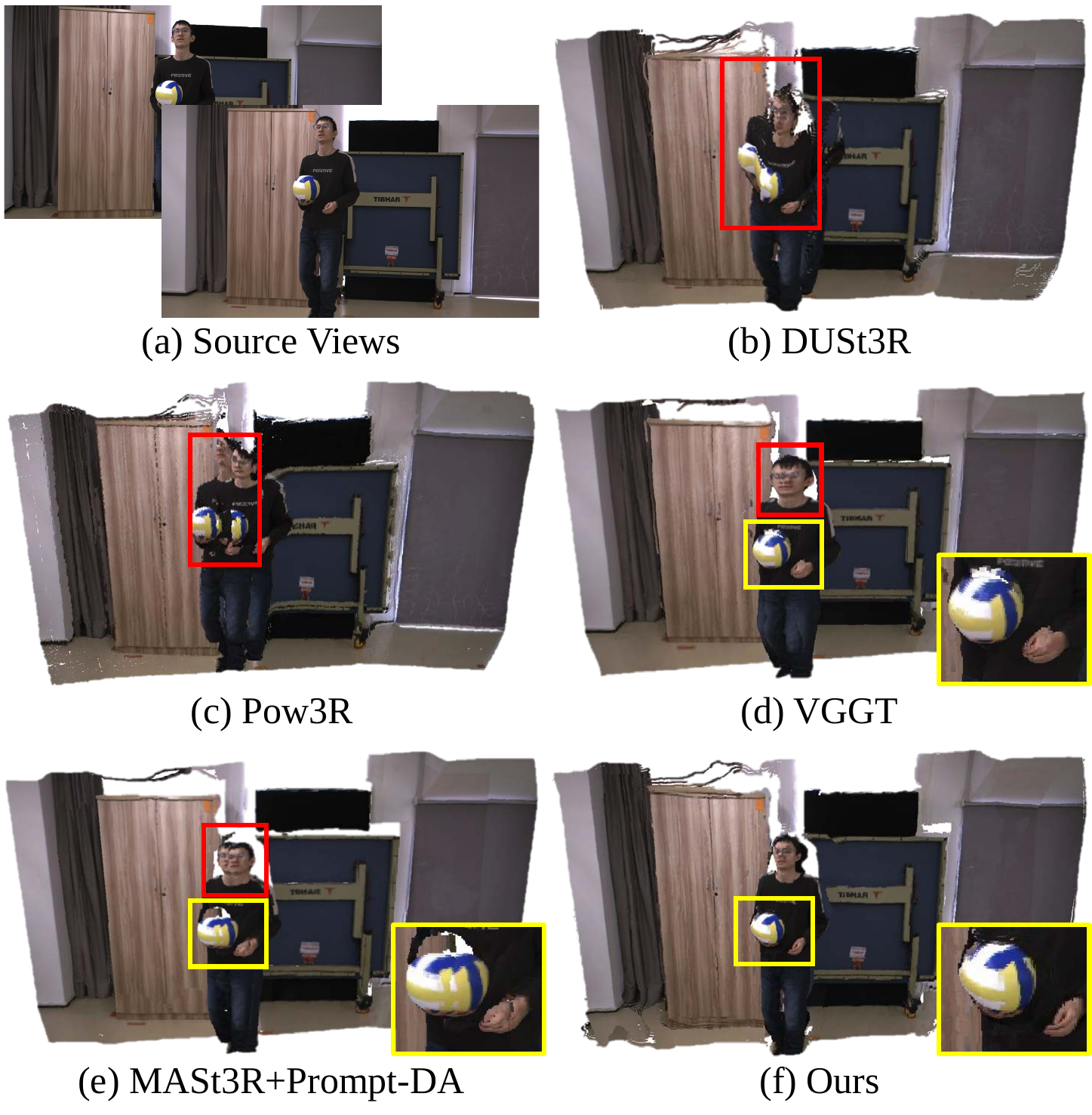}
  \vspace{-6mm}
  \caption{\textbf{Qualitative comparison on geometry.} From left to right, we show (a) source view images, the reconstruction point maps with corresponding pixels of (b) DUSt3R, (c) Pow3R, (d) VGGT, (e) MASt3R+Prompt-DA, and (f) Ours.}
  \label{fig:x_geo}
\end{figure}

\section{More Results on Geometry}

Our geometry module in stage 1 is able to generalize to different camera types, although our model is trained in a self-supervised manner without 3D loss.
In addition to the qualitative results on Mobile data in Fig.~\ref{fig:geo_ablation}, we illustrate the robustness of our geometry module in Fig.~\ref{fig:x_geo} on Industry Camera data.
DUSt3R~\cite{wang2024dust3r} and Pow3R~\cite{jang2025pow3r} struggle with large misalignment between two point maps on the foreground in Fig.~\ref{fig:x_geo}(b,c), even if Pow3R requires camera calibration.
Under only 2 sparse input views, VGGT~\cite{wang2025vggt} also badly aligns the foreground part of the point map, see face, ball and trousers in Fig.~\ref{fig:x_geo}(d).
By feeding MASt3R~\cite{leroy2024grounding} geometry as input, Prompt-DA~\cite{lin2025prompting} manages to refine the geometry of MASt3R but in a view-independent manner.
Thus the global alignment of two source views is not held.
Our affine transform module preserves the consistency of the overlapped part of two source views by using rendering loss and geometry regularization. 

\begin{figure}[h!]
  \centering
  \includegraphics[width=\linewidth]{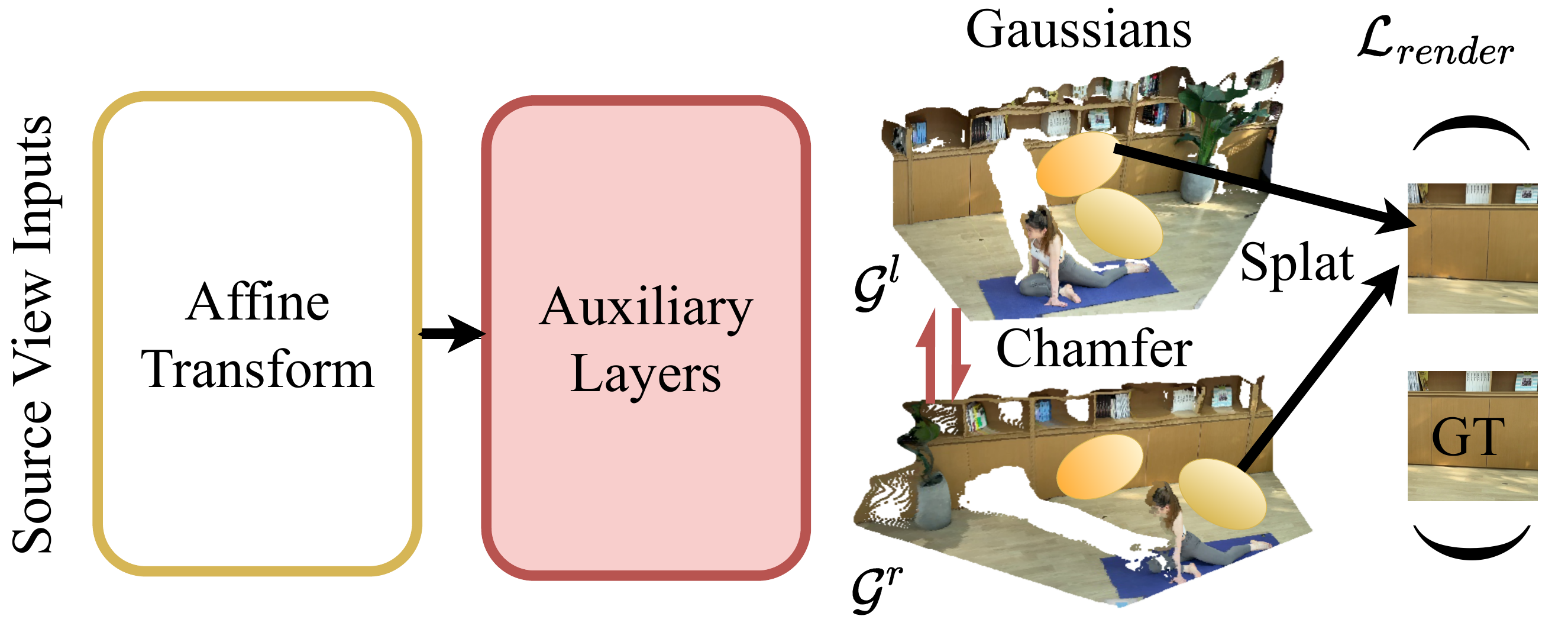}
  \caption{\textbf{Self-supervised training of stage 1.} We train the affine transform module and auxiliary layers by using Chamfer distance and rendering loss.}
  \label{fig:stage1}
\end{figure}

\begin{table}[t!]
\small
\centering
\begin{tabular}{l|ccc}
\toprule

\multicolumn{1}{c}{\multirow{1}[0]{*}{Method}} & PSNR$\uparrow$ & SSIM$\uparrow$ & LPIPS$\downarrow$ \\ \midrule

Stage 1 Render          & 24.844 & 0.794 & 0.296 \\
Stage 2 Initial Color          & 27.308 & 0.856 & 0.169 \\
Stage 2 Final Splatting         & \textbf{28.703} & \textbf{0.889} & \textbf{0.169} \\

\bottomrule
\end{tabular}
\caption{\textbf{Ablation study of rendering module.} We average the metrics across all datasets.}
\label{tab:render}
\end{table}

\begin{figure}[h!]
  \centering
  \includegraphics[width=1.\linewidth]{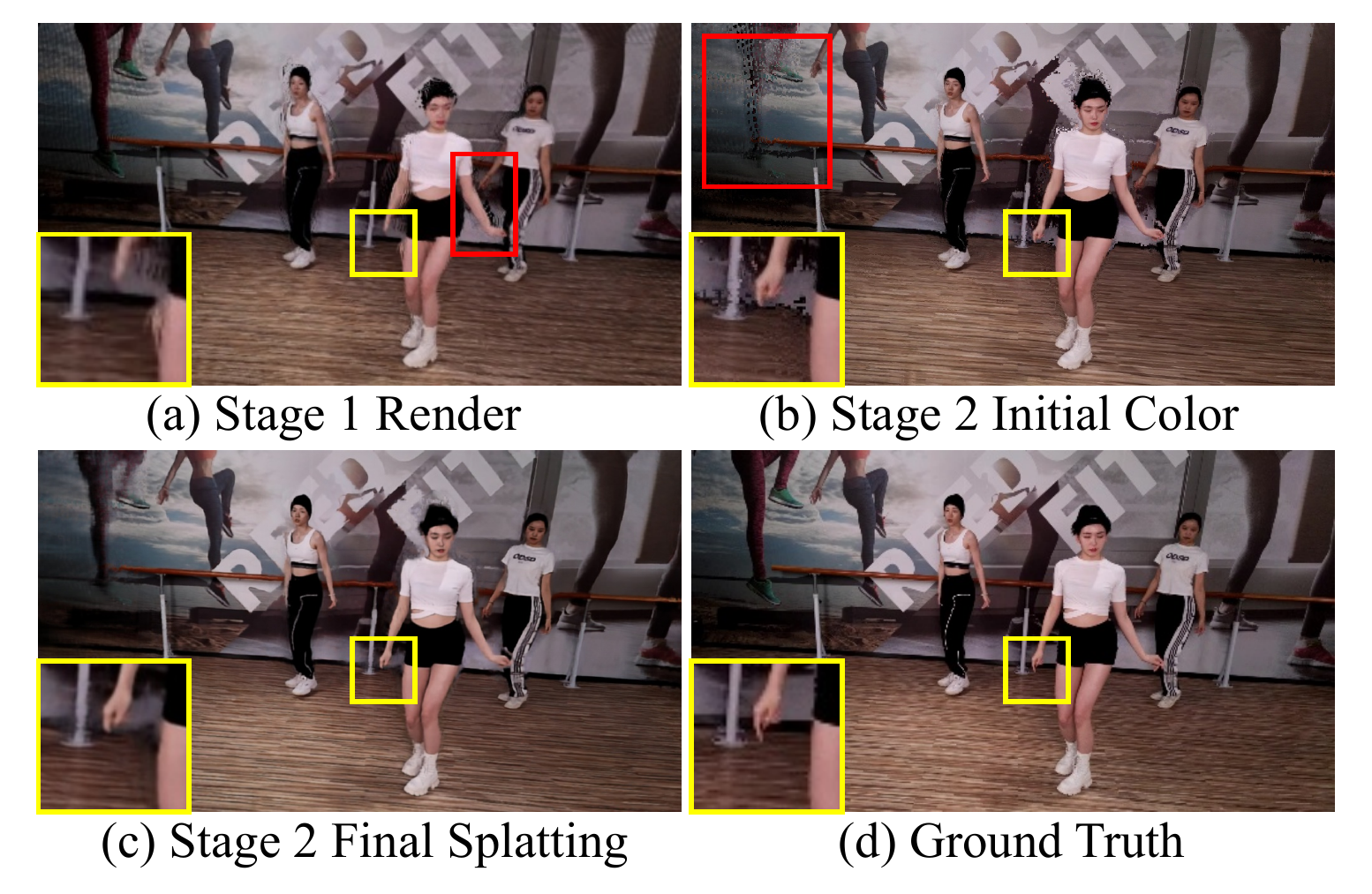}
  \vspace{-6mm}
  \caption{\textbf{Qualitative ablation results.} We show novel view synthesis results of (a) stage 1, (b) initial warping color of stage 2, (c) final splatting of stage 2, and (d) ground truth.}
  \label{fig:render_ablation}
\end{figure}

\section{Ablation Study on Rendering}
As mentioned in the Method section, our stage 1 network is associated with some auxiliary layers to generate two Gaussian planes, in order to supervise the geometry with rendering loss in a self-supervised manner, see Fig.~\ref{fig:stage1}.
However, such rendering typically struggles with some missing parts, Fig.~\ref{fig:render_ablation}(a) and obtains a low numeric result in Tab.~\ref{tab:render}, because the coarse registration in stage 1 can not totally handle the large sparsity of input views. 
After geometry refinement in stage 2, we can render the target view by warping the color from source views as in Eq.~\ref{eq:color_stage2}.  
However, the floating points between foreground and background would make noisy results, Fig.~\ref{fig:render_ablation}(b). 
Therefore, we use the color residual map in Eq.~\ref{eq:color} and Splatting mechanism to correct the artifacts in Fig~\ref{fig:render_ablation}(c), and to improve the quantitative result in Tab.~\ref{tab:render}. 

\begin{table}[t!]

\small
\centering
\begin{tabular}{c|l|c c}
\toprule

\multicolumn{2}{c}{\multirow{1}[0]{*}{Module}} & Time (ms) & Inp. Res.\\
\midrule
\multirow{2}[5]{*}{\rotatebox{90}{\scriptsize{Stage 1}}}
& Point Init. (MASt3R)          & 97 & \multirow{2}[5]{*}{\rotatebox{90}{\tiny{$512 \times 288$}}}  \\
& Affine Transform             & 34 & \\
& Depth Init.               & 5  & \\
\midrule
\multirow{3}[5]{*}{\rotatebox{90}{\scriptsize{Stage 2}}}& Depth Refine.             & 119 & \multirow{3}[5]{*}{\rotatebox{90}{\scriptsize{$1024 \times 576$}}} \\
& Color Init.                  & 5  & \\
& Gaussian Plane/Color Correct & 18 &  \\
& Splatting                    & 2  &  \\
\midrule
& Total      & 280 & \\

\bottomrule
\end{tabular}
\caption{\textbf{Time cost of our pipeline.} In stage 1, our network takes a pair of images in the resolution of $512 \times 288$ as inputs, while two images of $1024 \times 576$ are fed into our refinement module in stage 2. Our full pipeline takes totally 280ms.}
\label{tab:time_analysis}
\end{table}

\section{Time Analysis}
We conduct experiments on a machine equipped with an RTX 3090 GPU with 24GB memory for our method and provide a time analysis of our pipeline in Tab.~\ref{tab:time_analysis}.
In stage 1, point map reconstruction takes a lot of time by using original MASt3R~\cite{leroy2024grounding}, due to the complex structure of ViT, while our iterative affine transform is very efficient.  
In addition, the depth initialization is achieved by using $\alpha$-blending~\cite{kerbl2023_3dgs} with a fixed radius and diagonal rotation matrix.
Due to the fine resolution input in stage 2, the majority of time is used for depth refinement by using the costly 3D convolution.
The time analysis is done by using PyTorch and we believe that the full pipeline can be largely accelerated with a C++ implementation of TensorRT.

\section{Data} 

As mentioned in the Experiment section, we train and evaluate our method on 3 types of camera data, including industry camera (THumanMV~\cite{zhou2024gps}), GoPro, and mobile phone (4K4D~\cite{xu20244k4d} and SelfCap~\cite{xu2024representing}). 
For industry camera, we take 15/11 training/validation sequences in different scenes.
The validation data is, in general, unseen character or unseen motion.
For mobile stage data, we take around 3000 frames from 4K4D dance sequence and SelfCap yoga sequence for training and validate on the rest of the frames.
Since the industry camera and the mobile phone stage are typically complex capture systems with relatively large focal lengths, they can only capture small amplitude movement.
Therefore, we capture multi-person movement of sport in large scale scenes with a portable GoPro system in the mode of 1080P 30FPS.
For GoPro data, we take 6/4 sequences as training/validation data.
All aforementioned datasets provide camera calibration.
The majority of the data we used has been already publicly available for research purposes, while a part of the data is not available due to a confidential issue.

\end{document}